\documentclass[letterpaper]{article} 
\usepackage{aaai24}  
\usepackage{times}  
\usepackage{helvet}  
\usepackage{courier}  
\usepackage[hyphens]{url}  
\usepackage{graphicx} 
\usepackage{natbib}  
\usepackage{caption} 
\usepackage{algorithm}
\usepackage{algorithmic}

\usepackage{multirow}
\usepackage{makecell}
\usepackage{newfloat}
\usepackage{listings}

\usepackage{enumitem}
\usepackage{tikz}
\usepackage{amsmath}
\usepackage{booktabs}
\usepackage{lipsum}
\usepackage{listings}
\usepackage{epsfig,endnotes}
\usepackage{graphicx}
\usepackage{array, multirow}
\usepackage{makecell}
\usepackage{tabularx}
\usepackage{xspace}
\usepackage{array}
\usepackage{caption}
\usepackage{glossaries}
\usepackage{varwidth}
\usepackage{xcolor}
\usepackage{colortbl}
\usepackage{subcaption}

\usepackage{mathtools} 
\usepackage{amsfonts} 
\usepackage{verbatim}
\usepackage{framed}
\usepackage{inconsolata}
\usepackage{fancyvrb}
\usepackage{amsthm}
\usepackage{dblfloatfix}

\newtheorem{finding}{\textit{Finding}}

\newcommand{\para}[1]{\noindent\paragraph{#1}\quad}

\usepackage{newfloat}
\usepackage{listings}
\DeclareCaptionStyle{ruled}{labelfont=normalfont,labelsep=colon,strut=off} 
\floatstyle{ruled}
\newfloat{listing}{tb}{lst}{}
\floatname{listing}{Listing}
\title{Can LLM Replace Stack Overflow? A Study on Robustness and Reliability of Large Language Model Code Generation}
\author{
    Li Zhong,
    Zilong Wang
}
\affiliations{
    University of California, San Diego\\
    lizhong@ucsd.edu, zlwang@ucsd.edu

}

\newcommand{\dataset}{\textsc{RobustAPI}\xspace}

\begin{document}

\maketitle

\begin{abstract}
Recently, large language models (LLMs) have shown an extraordinary ability to understand natural language and generate programming code. It has been a common practice for software engineers to consult LLMs when encountering coding questions. Although efforts have been made to avoid syntax errors and align the code with the intended semantics, the reliability, and robustness of the code generation from LLMs have not yet been thoroughly studied. The executable code is not equivalent to reliable and robust code, especially in the context of real-world software development. For example, the misuse of APIs in the generated code could lead to severe problems, such as resource leaks, program crashes, etc. Existing code evaluation benchmarks and datasets focus on crafting small tasks such as programming questions in coding interviews. However, this deviates from the problems developers typically consult LLMs about. To fill the missing piece, we propose a dataset \dataset{} for evaluating the reliability and robustness of code generated by LLMs. We collect 1208 coding questions from Stack Overflow on 18 representative Java APIs. We summarize the common misuse patterns of these APIs and evaluate them on current popular LLMs. The evaluation results show that even GPT-4 has 62\% of the generated code that contains API misuses. It would cause unexpected consequences if the code is introduced into real-world software.
\end{abstract}

\section{Introduction}


The new era of language modeling arrives when large language models (LLMs) are capable of generating customized code according to the user's needs~\cite{ye2023comprehensive,OpenAI2023GPT4TR,anil2023palm}. It is not surprising that more and more software engineers choose to query large language models for the answer to the coding questions, such as generating a code snippet using certain APIs or detecting bugs in a few lines of code. Large language models are able to respond more suitable and customized answers for the question compared with searching in the online programming forums, such as Stack Overflow.

\begin{figure*}[t]
    \centering
    \includegraphics[width=0.95\linewidth]{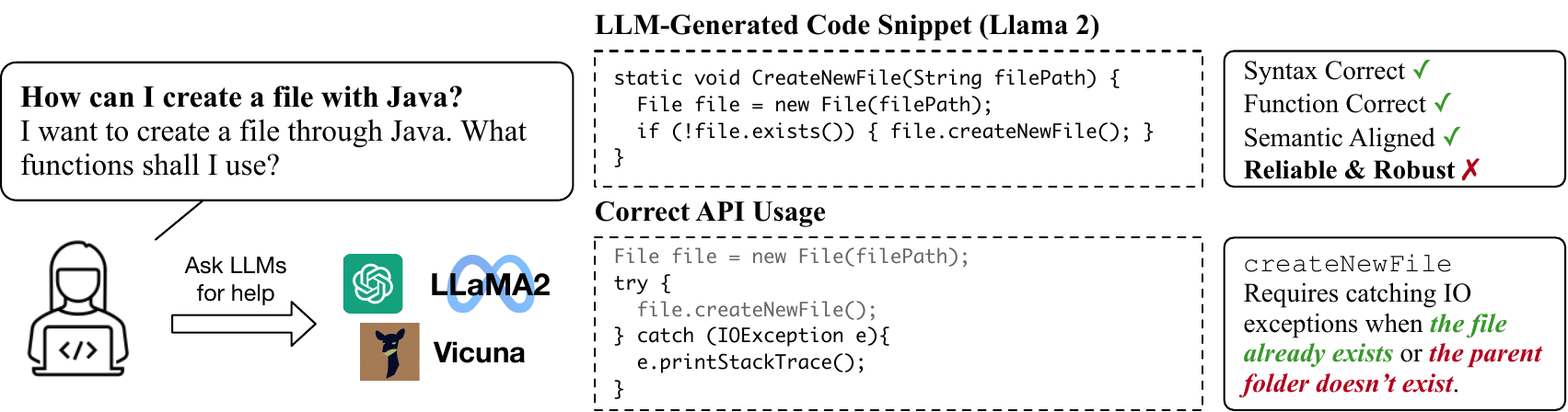}
    \caption{The scenario where software engineers consult large language models for the answer to the programming
questions. The generated code snippet is not reliable and has potential risks in the software development.}
    \label{fig:misuse-example}
\end{figure*}

Such a fast pace conceals potential risks in the code generation of large language models. From the perspective of software engineering, the robustness and reliability of generated code have not yet been thoroughly studied even if numerous works have been made to avoid syntax errors and improve semantic understanding in the generated code~\cite{xu2022systematic,chen2021evaluating,shen2023pangu,luo2023wizardcoder}. Unlike the online programming forums, the generated code snippets are not reviewed by the community peers and thus suffer from API misuse, such as missing boundary checking in file reading and variable indexing, missing file stream closing, failure in transaction completion, etc. Even if the code samples are executable or functionally correct, misuse can trigger serious potential risks in production, such as memory leaks, program crashes, garbage collection failures, etc, as shown in Figure \ref{fig:misuse-example}. To make things worse, the programmers asking these questions could be vulnerable to the risk if they are novices to the APIs and cannot tell the violations in the generated code snippets. Therefore, it is essential to contemplate the code reliability while evaluating the code generation by large language models.

To evaluate the code generation of large language models, most of the existing benchmarks focus on the functional correctness of the execution result from the generated code, which means the code is acceptable as long as it is functional for the user's purpose~\cite{chen2021evaluating,yin2018learning,lu2021codexglue}. We argue that the correct execution result is important but it is not only the case in the software development scenario. What the engineers really need is a reliable code sample without potential risks in the long run. Moreover, the domain of most current programming datasets is far from software engineering. The data source is mostly online coding challenge websites, such as Codeforces, Kattis, Leetcode, etc~\cite{hendrycks2021measuring,austin2021program}. Although remarkable progress has been made, we argue that they fail to substantially help the software development in practical scenarios. 

To this end, we propose \dataset{}, a comprehensive benchmark to evaluate the reliability and robustness of code generated by large language models, including a dataset of coding questions and an evaluator using the abstract syntax tree (AST)~\cite{fischer2007abstract}. 
In the dataset, we target creating an evaluation setting that is close to real software development. Thus we collect representative questions about Java from Stack Overflow. Java is one of the most popular programming languages and is widely used in software development because of its \textit{write once, run anywhere (WORA)} feature\footnote{\url{https://en.wikipedia.org/wiki/Java_(programming_language)}}. For each question, we provide a detailed description and the related Java API. We design templates to trigger large language models to generate the code snippet and the corresponding explanation. 
We also provide an evaluator that analyzes the generated code snippets using the abstract syntax tree (AST) and compares them with the expected API usage patterns. Following \citet{zhang2018code}, we formalize the API usage patterns into structured call sequences, as shown in Figure \ref{fig:api_checker}. The structured call sequences present how these APIs can be properly used to eliminate the potential system risks. Any violations of such structured call sequences would be considered as API misuse from the perspective of software engineering. 

We collect 1208 real questions from Stack Overflow which involves 18 representative Java APIs. We run experiments on the close-sourced language models (GPT-3.5 and GPT-4~\cite{OpenAI2023GPT4TR}) as well as the open-sourced language models (Llama-2~\cite{touvron2023Llama}, Vicuna-1.5~\cite{vicuna2023}. We use the default hyper-parameter settings of the models without extensive hyper-parameter tuning. We further design two experiment settings, zero-shot and one-shot, where none or one demonstration sample is provided in the prompt. We conduct a comprehensive analysis of the generated code and study the common API misuse cases of current large language models. We would like to bring up the important issues of API misuse in the code generation by large language models, and provide a new dimension to evaluate large language models other than the commonly-used functional correctness. The main purpose of this benchmark is not to evaluate the functional correctness of the generated code, but instead, we focus on reliability and robustness. We hope this work could facilitate future research on this topic and help create a more robust coding helper out of large language models to step further into real artificial general intelligence. We open-source our dataset and evaluator on GitHub\footnote{https://github.com/FloridSleeves/RobustAPI}. We summarize our contribution as follows.
\begin{itemize}
    \item We propose a new benchmark, \dataset{}, to evaluate the reliability and robustness of code generation by large language models. This is an important but not yet well-studied perspective to evaluate the code quality apart from functional correctness.
    \item We provide a well-formalized evaluation framework including a dataset of Stack Overflow questions and an API usage checker using AST. We report the performance of popular large language models, including GPT-3.5, GPT-4, Llama-2, and Vicuna-1.5.
    \item We conduct a comprehensive analysis of the code generation performance of current large language models. We summarize the common API misuse for each model and point out the promising improvement direction for the future research.
\end{itemize}



\section{Related Work}

\para{Code Quality of LLM-Sythesized Code} With the release of Copilot~\cite{chen2021evaluating} and other commercial code assistant tools based on LLMs, the security and code quality of these tools gradually get the attention of the research community. \citet{yetistiren2022assessing} assess the quality of LLM-generated code from the aspects of compilation correctness, functional correctness, and code efficiency. \citet{siddiq2022empirical} studied code smells in code generated by LLMs, which is the poor design in code like unusually long method, or duplicated code. \citet{poesia2022synchromesh} shows that LLMs can make implementation errors in the code like syntax errors or semantic errors deviating from users' intention. \citet{jesse2023large} studied simple, stupid bugs in Codex and other LLMs, which shows that AI code assistants can help avoid some of such simple bugs but have a higher chance of introducing bugs that are hard to detect. As for security impact, \citet{pearce2022asleep} designed 89 security-sensitive scenarios for Copilot to complete the code for users, which shows approximately 40\% of the code is vulnerable. \citet{perry2022users} conducted the first large-scale user study to examine whether users interacting with AI Code assistants write secure code. They find that those users wrote significantly less secure code while they believe their code was secure. \citet{sandoval2023lost} conducts a user study to assess the security of low-level code with pointer and array manipulations generated by AI-based coding assistants. They find under this specific scenario, the assistants do not introduce more security bugs than humans. \citet{liu2023your} enlarges HumanEval~\cite{chen2021evaluating} by generating test cases with higher coverage which serve as an add-on to the existing programming benchmarks but the evaluation still focuses on functional correctness and simple programming questions far from software development. 
\citet{shen2023chatgpt} evaluates the reliability of ChatGPT by testing on adversarial examples, which however has a different meaning of `reliability' in their context. In this paper, we refer to reliability as the ability of code to resist failure, high workload, and unexpected input. 

\para{Quality Assessment of Code in Online Forum} Existing literature in the software engineering field has investigated the quality of code from online forums and warned developers of the potential issues. \citet{yang2016query} finds that the majority of code examples given in Stack Overflow answers cannot be compiled. \citet{zhou2016api} pointed out that 43\% of the posts investigated by them contained deprecated APIs, while \citet{fischer2017stack} found that 29\% of the code contains security risks. In \citet{zhang2018code}, the authors analyze the code by call sequence extraction and slicing, and compare it to the manually validated API usage rules, which concludes that 31\% of the code examples in Stack Overflow answers contain API misuse and could produce unexpected behaviors.

\section{Methodology}

In this section, we describe \dataset{}, a comprehensive benchmark to thoroughly evaluate the reliability and robustness of LLM-generated code. We describe the process of data collection and prompt generation when constructing the dataset. Then we present the API misuse patterns evaluated in \dataset{} and discuss the potential consequence of violations. Finally, we introduce the static analysis method in \dataset{} for detecting the API usage violations which leverages the abstract syntax tree and achieves higher evaluation accuracy in evaluating the API misuse in code generated by LLMs compared to rule-based method such as keywords matching.

\subsection{Data Collection}
To take advantage of the existing research efforts in the software engineering field, we build \dataset{} based on the dataset from ExampleCheck~\cite{zhang2018code} as our starting point. ExampleCheck is proposed to study the frequent Java API misuse in online Q\&A forums. We select 18 popular Java APIs from the dataset as shown in Table~\ref{tab:java_api_misuse}. These 18 APIs cover 6 domains including string processing, data structure, mobile development, crypto, I/O and database operation. Then we crawl questions relevant to these APIs from Stack Overflow. We only select the questions with online answers and we keep the questions whose provided answer contains API misuse. In this way, we guarantee that the questions in \dataset{} are answerable and non-trivial so we can use them to effectively evaluate the LLMs' ability in answering coding questions that \textit{humans are prone to make mistakes}. After filtering, we get 1208 questions in total. The distribution of questions for each domain is shown in Table~\ref{tab:java_api_misuse}.

\begin{table}[!hbt]

    \centering
    \resizebox{1.01\linewidth}{!}{
    \setlength{\tabcolsep}{0.4mm}{
    \begin{tabular}{lccc}
        \toprule
        \bf API & \bf Domain & \bf Conseq* & \bf Github*\\
        \hline
        StringTokenizer.nextToken & \multirow{3}{*}{\makecell{String\\Process\\ (307)}}  & (iii) & 13.3K\\
        String.getBytes & & (iii) & 88.1K\\
        JsonElement.getAsString &  & (iii) & 4.4K\\
        \hline
        List.get & \multirow{3}{*}{\makecell{Data \\Structure\\ (404)}} & (iii) & 2.7M\\
        Map.get &  & (iii) & 2.4M\\ 
        Iterator.next & & (iii) & 918K\\
        \hline
        ProgressDialog.dismiss & \multirow{4}{*}{\makecell{Mobile \\ Develop\\(75)}} & (iii) & 54K\\
        TypedArray.getString &  & (iv) & 6.8K\\
        ApplicationInfo.loadIcon & & (v) & 3.6K\\
        Activity.setContentView & & (v) & 4.6K\\
        \hline
        Cipher.init & \multirow{1}{*}{Crypto (10)} & (iii)& 66.3K\\ 
        \hline
        RandomAccessFile.write & \multirow{6}{*}{I/O (390)} & (i)& 129K\\
        BufferedReader.readLine & & (iii) & 74.8K\\
        PrintWriter.write & & (i) & 1.1M\\
        File.mkdirs & & (ii) & 73.2K\\
        File.createNewFile & & (i) & 176K\\
        FileChannel.write & & (i)& 5.2K\\
        \hline
        SQLiteDatabase.query & Database (22) & (iv) & 4K\\
        \midrule
        \textbf{Total} & 1208 & & 7.8M\\
        \bottomrule
    \end{tabular}
    }
    }
    \caption{18 popular Java APIs in \dataset{}. They are easily misused by developers according to the existing literature of software engineering~\cite{zhang2018code}. *Consequences: (i) data loss; (ii) file system corruption; (iii) program crash; (iv) resource leak; (v) user interface bug. *Github: occurrences of this API on Github.}
    \label{tab:java_api_misuse}
\end{table}

After collecting the questions, we convert them into the JSON format with the following fields: \texttt{\{id, api, question, origin\}}. \texttt{id} field contains the unique id we assign for each sample. \texttt{api} field contains the API that we specifically instruct the large language models to use as a question hint. \texttt{question} field contains the title and description of the Stack Overflow questions. \texttt{origin} field contains the original URL of this sample. 

\subsection{Prompt Generation} 
In the prompt, we start with the task introduction and the required response format. Then we append the few-shot demonstrations on this API when conducting experiments in the few-shot settings. The demonstration examples satisfy our provided response format. Next, we append the question and the corresponding API hint for this question. This prompt simulates a user asking coding questions without providing any additional hints from the API documentation which is a typical scenario when novice developers seek help from large language models. Due to the chat completion nature of state-of-the-art LLMs, we wrap the question and answer with special tags to instruct LLMs to generate answers to the questions. The prompt template is adapted from \cite{patil2023gorilla}, which can help LLMs follow a specific generation template so that we can extract more compilable code snippets from the response. 

\subsection{Demonstration Samples}\label{sec:demo}
Demonstration samples have been proven helpful to LLMs in understanding natural language. To thoroughly analyze LLMs' ability in code generation, we design two few-shot settings, One-shot-irrelevant and One-shot-relevant. 

In the one-shot-irrelevant setting, we provide LLMs with an example using an irrelevant API (e.g. \texttt{Arrays.stream}). We assume this demonstration example would eliminate the syntax errors in the generated code. 

In the one-shot-relevant setting, we provide LLMs with an example using the same API as the given question. The provided example contains a pair of question and answer. The question in the demo example is not present in the testing dataset and we manually revise the answer to ensure that there is no API misuse in it and that the semantics well align with the questions.


\subsection{Java API Misuse}
\begin{figure*}[!htb]
    \centering
    \includegraphics[width=0.9\linewidth]{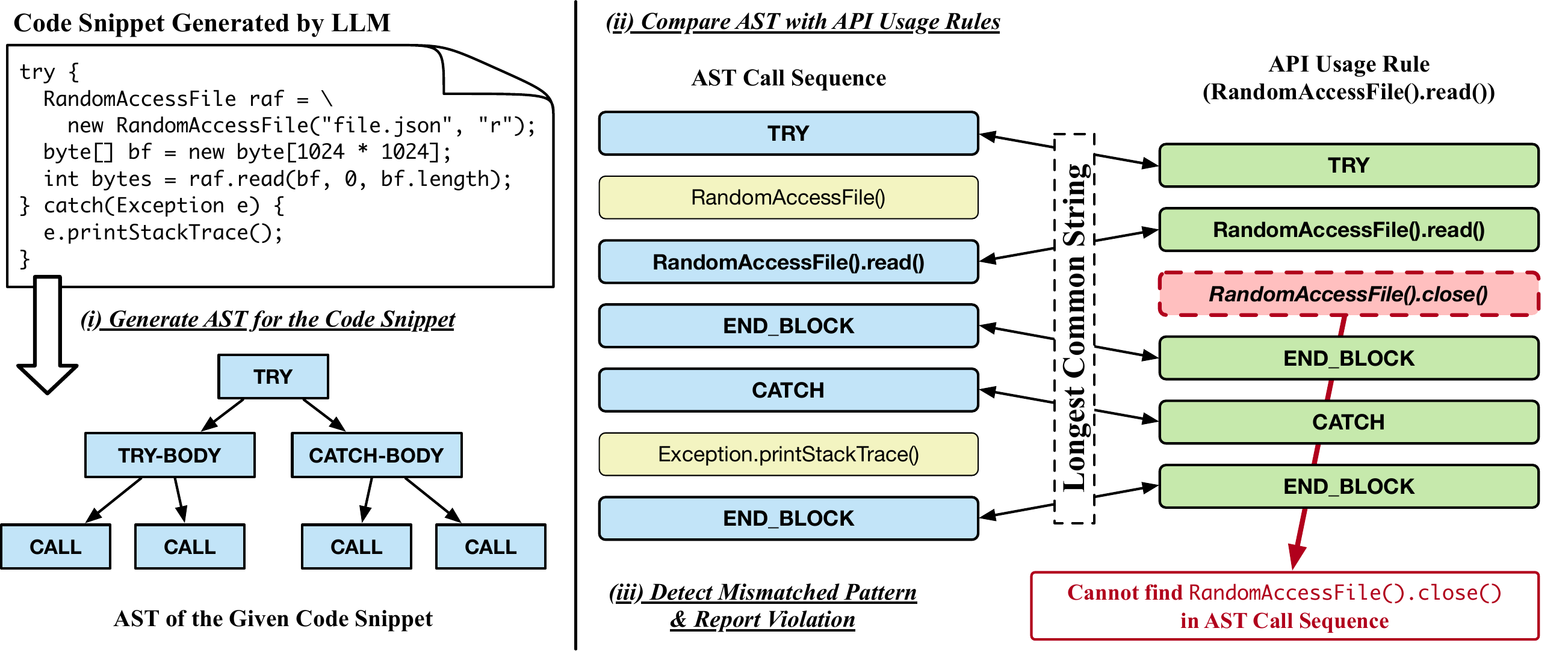}
    \caption{The workflow of Our API Checker. The API checker uses the static analysis method and analyzes the generated code with the abstract syntax tree (AST). The API misuse is detected when the AST call sequence and the API usage rule do not match.}
    \label{fig:api_checker}
\end{figure*}
When using the APIs provided by language libraries, developers need to follow the API usage rules so that they can take full advantage of the ideal API effect. Violating these rules and misusing the APIs could result in unexpected behaviors in production. A typical example is the file operation. When opening and writing a file through \texttt{RandomAccessFile}, two usage rules need to be enforced: (1) Reading the file could throw exceptions. If the buffer limit is reached before the expected bytes are read, the API would throw \texttt{IndexOutOfBoundsException}. Also, if the file is concurrently closed by other processes, the API would throw \texttt{ClosedChannelException}. To deal with these exceptions, the correct implementation should enclose the API inside \texttt{try-catch} blocks. (2) The file channel should be closed after usage. Otherwise, if this code snippet is inside a long-lasting program that is concurrently running in multiple instances, the file resources could be run out. Therefore, the code needs to invoke \texttt{close} API after all file operations. The correct usage are shown as following:

\begin{Verbatim}[frame=single,fontfamily=cmss,commandchars=\\\{\},fontsize=\scriptsize]
\textbf{Correct API Usage:}
\texttt{try \{}
    \texttt{RandomAccessFile raf =} 
    \texttt{  new RandomAccessFile("/tmp/file.json", "r");}
    \texttt{byte[] buffer = new byte[1024 * 1024];}
    \texttt{int bytesRead = raf.read(buffer, 0, buffer.length);}
    \texttt{raf.close();}
\texttt{\} catch(Exception e) \{...\}}
\end{Verbatim}
In \dataset{}, we summarized 41 API usage rules from the 18 APIs, which are validated in the documentation of these APIs~\cite{zhang2018code}. These rules include: (1) The guard condition of an API, which should be checked before API calls. For example, check the result of \texttt{File.exists()} before \texttt{File.createNewFile()} (2) Required call sequence of an API, which should be called in a specific order. For example, call \texttt{close()} after \texttt{File.write()}. (3) Control structures of an API. For example, enclose \texttt{SimpleDateFormat.parse()} with \texttt{try-catch} structure. 

\subsection{Detecting API Misuse}

Existing research in evaluating the code generated by LLMs usually uses test cases, which falls short when testing the reliability and robustness of code.
To deal with this challenging problem, we use static analysis for \dataset{}, which has relatively mature solutions in detecting API misuse~\cite{zhang2018code,nguyen2014mining,wang2013mining, huang2023protecting}. To evaluate the API usage correctness in code, \dataset{} detects the API misuses against the API usage rules by extracting call consequences and control structures from the source code, as shown in Figure~\ref{fig:api_checker}. 
The code checker first checks the code snippets to see whether it is a snippet of a method or a method of a class so that it can enclose this code snippet and construct an abstract syntax tree (AST) from the code snippet. Then the checker traverses the AST to record all the method calls and control structures in order, which generates a call sequence. Next, the checker compares the call sequence against the API usage rules. It infers the instance type of each method call and uses the type and method as keys to retrieve corresponding API usage rules. Finally, the checker computes the longest common sequence between the call sequence and the API usage rules. If the call sequence does not match the expected API usage rules, the checker will report API misuse.

\section{Experimenet}

\subsection{Experiment Setup}
In the experiments, we evaluate \dataset{} on four LLMs: 
GPT-3.5~\cite{OpenAI2023GPT4TR}, GPT-4~\cite{OpenAI2023GPT4TR}, Llama-2~\cite{touvron2023Llama}, Vicuna-1.5~\cite{vicuna2023}.
We use the default hyper-parameter settings of each model without further extensive hyper-parameter tuning. All experiment results are Pass@1 unless specified.
For all models, we evaluate three experiment settings:
\begin{itemize}
    \item \textbf{Zero-shot:} No example is provided in the prompt. The prompt only contains the instruction, question.
    \item \textbf{One-shot-irrelevant:} \dataset{} provides one example of an irrelevant task in the prompt.
    \item \textbf{One-shot-relevant:} \dataset{} provides one example of the same API with the correct usage in the prompt.
\end{itemize}
The examples for shot generations are manually written and double-checked by the authors. Then they are evaluated against the API usage checkers to make sure they are aligned with the API usage rules.

\subsection{Evaluation Metrics}

To quantitatively evaluate the reliability of the generated code, we define the following values and our metrics are computed based on them. Supposing that we have $N$ questions in our dataset, we divide them into three groups.
\begin{itemize}
    \item $N_{\text{misuse}}$: The number of cases where our API usage checker detects the API usage violations.
    \item $N_{\text{pass}}$: The number of cases where our API usage checker does not detect the API usage violations.
    \item $N_{\text{non-comp}}$: The number of cases where the LLM fails to generate code or the generated code is not compilable.
\end{itemize}
Based on the values, we define our metrics.
\begin{itemize}
    \item \textbf{API Misuse Rate $=N_{\text{misuse}} / (N_{\text{misuse}}+N_{\text{pass}})$:} To analyze the proportion of misuse cases among the compilable code snippets. It reveals how reliable the generated code is after the users filter out the non-compilable cases.
    \item \textbf{Compilation Rate $=(N_{\text{misuse}}+N_{\text{pass}})/N$:}
    To analyze the proportion of compilable cases among all questions. It is necessary to consider the percentage of compilable cases in order to eliminate the influence from the extreme situations, such as when only a few compilable code snippets are generated.
    \item \textbf{Overall API Misuse Percentage $=N_{\text{misuse}}/N$:} To analyze the proportion of misuse cases among all questions. 
\end{itemize}

\subsection{Research Questions}
We conduct a series of experiments on state-of-the-art LLMs based on \dataset{}, which demonstrate the usability and effectiveness of \dataset{}. The experiments provide insights on the ability to answer real-world coding questions and the robustness and reliability of these answers regarding API misuse problems. In the experiment, we try to answer the following questions:
\begin{itemize}
    \item \textbf{Q1:} What are the API misuse rates in answering real-world coding questions by these LLMs?
    \item \textbf{Q2:} How do irrelevant shots affect the results?
    \item \textbf{Q3:} Can correct API usage examples reduce the misuse?
    \item \textbf{Q4:} Why does LLM-generated code fail the API usage check?
\end{itemize}

\begin{table*}[!htbp]
  \centering
  \small
\resizebox{\linewidth}{!}{
\begin{tabular}{lccccccccc}
\toprule
\multirow{3}[4]{*}{\bf Model} & \multicolumn{3}{c}{\textbf{Zero-shot}} & \multicolumn{3}{c}{\textbf{One-shot-irrelevant}} & \multicolumn{3}{c}{\textbf{One-shot-relevant}} \\
\cmidrule{2-10}      & \textbf{Misuse} & \textbf{Compilable} & \textbf{Overall } & \textbf{Misuse} & \textbf{Compilable} & \textbf{Overall } & \textbf{Misuse} & \textbf{Compilable} & \textbf{Overall } \\
      & \textbf{Rate $\downarrow$} & \textbf{Rate $\uparrow$} & \textbf{Misuse $\downarrow$} & \textbf{Rate $\downarrow$} & \textbf{Rate $\uparrow$} & \textbf{Misuse $\downarrow$} & \textbf{Rate $\downarrow$} & \textbf{Rate $\uparrow$} & \textbf{Misuse $\downarrow$} \\
\midrule
\textbf{GPT 3.5} & 62.97\% & 79.14\% & 49.83\% & 68.09\% & 91.06\% & 62.00\% & 38.56\% & 80.71\% & 31.13\% \\
\textbf{GPT 4} & 68.81\% & 90.23\% & 62.09\% & 70.38\% & 91.39\% & 64.32\% & 54.40\% & 90.40\% & 49.17\% \\
\textbf{Llama 2}$^\ast$ & 7.34\%$^\ast$ & 9.02\%$^\ast$ & 0.66\%$^\ast$ & 61.36\% & 80.13\% & 49.17\% & 64.47\% & 72.93\% & 47.02\% \\
\textbf{Vicuna 1.5} & 45.66\% & 37.17\% & 16.97\% & 57.85\% & 83.86\% & 48.51\% & 42.53\% & 64.24\% & 27.32\% \\
\textbf{ds-coder-6.7b-base} & 41.55\% & 40.65\% & 16.89\% & 75.60\% & 95.90\% & 72.43\% & 64.12\% & 67.14\% & 43.05\% \\
    \textbf{ds-coder-6.7b-instruct} & 47.52\% & 50.00\% & 23.76\% & 59.04\% & 96.61\% & 57.04\% & 38.40\% & 86.01\% & 33.03\% \\
\bottomrule
\end{tabular}%
    }
  \caption{Performance of Each LLM on \dataset{}. $\downarrow$: the \textit{lower} the \textit{better}. $\uparrow$: the \textit{higher} the \textit{better}. Misuse Rate is the proportion of misuse cases among the compilable cases; Compilation Rate is the proportion of compilable cases among all questions; Overall Misuse is the proportion of misuse cases among all questions. $^\ast$Though Llama2 has a low misuse rate, its compilation rate is significantly lower than other models.}
  \label{tab:addlabel}%
\end{table*}%

\subsection{API Misuse Rate}
\begin{figure}[!htb]
    \centering
    \includegraphics[width=0.9\linewidth]{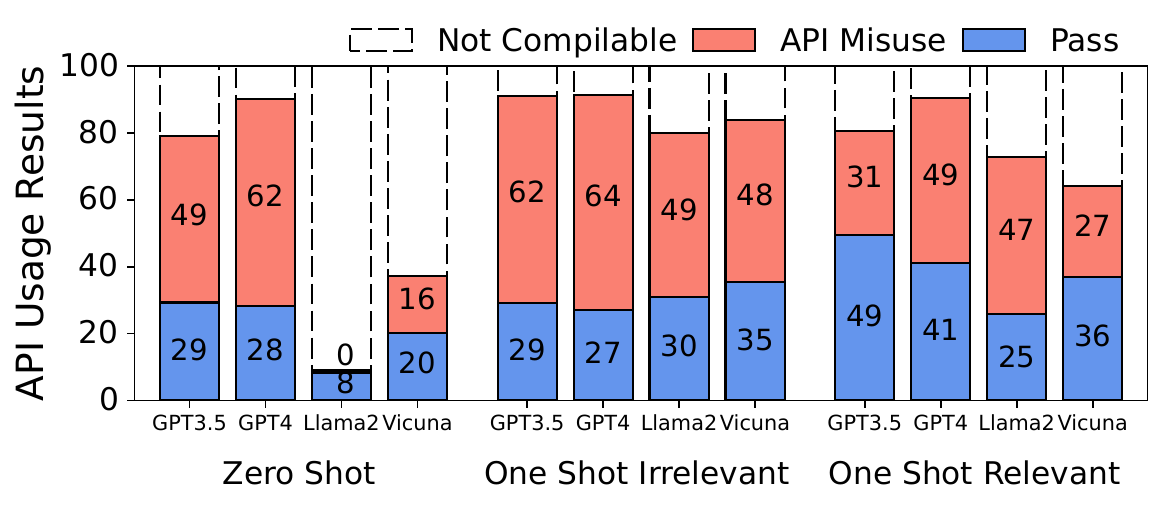}
    \caption{Result of Checking API Usage from LLMs. Red bars are the percentage of answers that contain API misuse, which is the \textit{lower}, the \textit{better}. The white bars in dot lines are the percentage of code answers that are not compilable.}
    \label{fig:api_misuse_rate}
\end{figure}
Firstly, we present the API misuse rate of each model based on \dataset{} on the left of Figure~\ref{fig:api_misuse_rate}. In this figure, the higher the API misuse rate is, the worse the code reliability and robustness for this large language model. The API misuse rate is calculated by dividing answers that can be compiled and contains API misuses by all the answers that can be compiled.
From the evaluation results, all the evaluated models suffer from API misuse problems, even the state-of-the-art commercial models like GPT-3.5 and GPT-4. In zero-shot settings, Llama has the lowest API misuse rate. However, this is partially due to that most of Llama's answers do not include any code. A counter-intuition finding is that GPT-4 actually has a higher API misuse rate than GPT-3.5, though the coding ability of GPT-4 is proved to be \textit{``40\% more advanced than its predecessor, GPT-3.5"}~\cite{openai2023gpt4}. We also evaluate a code-specialized large language model, DeekSeekCoder\cite{piplani2018deepseek}, which is trained on a variety of programming languages including Java, and surpasses many existing Code LLMs. We report the results of \texttt{deepseek-coder-6.7b-base} and \texttt{deepseek-coder-6.7b-instruct}. We observe that the code-specialized large language model can generate more compilable samples. However, the API misuse rate is not significantly better than other models. This indicates that with the code generation ability of large language models is largely improved nowadays, the reliability and robustness of code in real-world production rises as an unnoticed issue. And the space for improvement is huge for this problem. 

The execution time for static analysis is shown in Table~\ref{tab:exe-time}. The time difference is due to the different coding styles of each LLM, all of which are within 7 minutes.
\begin{table}[htbp]
  \centering
    \small
\resizebox{\linewidth}{!}{
    \setlength{\tabcolsep}{1mm}{
\begin{tabular}{ccccc}
\toprule
\textbf{GPT 3.5} & \textbf{GPT 4} & \textbf{Llama 2} & \textbf{Vicuna 1.5} & \textbf{DeepSeek-Coder} \\
\midrule
6m 31s & 6m 56s & 6m 36s & 6m 19s & 6m 36s \\
\bottomrule
\end{tabular}%
}}
  \caption{Execution Time of Static Analysis in \dataset{}.}
  \label{tab:exe-time}%
\end{table}%

\begin{finding}
Answers to real-world coding questions from the state-of-the-art large language models widely have API misuse problems.
\end{finding}    

\subsection{One-Shot-Irrelevant Results}
In this experiment, \dataset{} gives a pair of question and answer as an example to show the model how to follow the template required by the instructions. The example contains no information about the API usage checked by \dataset{}. The result is shown in the middle of Figure~\ref{fig:api_misuse_rate}. However, for most models, the irrelevant shot does not significantly reduce the API misuse rate but on the contrary, slightly increases the misuse rate. One possible reason for this is the irrelevant shot provided to the large language models actually encourages the models to give a lengthy code solution, which increases the chance of API misuse. API misuse rate of Llama increases significantly after adding the irrelevant shot because it has more valid answers that contain code snippets. Overall, adding an irrelevant shot triggers the large language models to generate more valid answers, which enables a better evaluation of the code reliability and robustness.

\begin{finding}
Among all the answers containing compilable code, 57-70\% of the LLM answers contain API misuse, which could lead to severe consequence in production.
\end{finding}

\begin{finding}
Irrelevant shot examples does not help decrease the API misuse rate but triggers more valid answers, which show to be effective for benchmarking the model performance.
\end{finding}

\subsection{One-Shot-Relevant Results}
In this experiment, \dataset{} adds a manually-written shot in the prompt, which performs a different task but uses the same API. This gives hints to LLMs on how to use these APIs correctly. From the results, after adding the correct usage shot, the API misuse rates of GPT-3.5, GPT-4, and Vicuna significantly drop. This indicates an effective improvement under this experiment setting. As for Llama, the relevant shot does not improve the performance. This experiment shows that some LLMs can effectively `learn' the correct API usage and follow the usage. However, since existing language models are trained with data from code repositories if the training datasets contain a large number of API violations, the language models are prone to generate code with API misuses, which explains the high API misuse rate in zero-shot and one-shot-irrelevant evaluation. We show Pass@k results of one-shot-relevant in Table~\ref{tab:pass-k}.
\begin{table}[htbp]
  \centering
      \small
    \setlength{\tabcolsep}{1mm}{
    \begin{tabular}{lccc}
    \toprule
    \bf Pass@k & \textbf{Misuse Rate} & \textbf{Compilation Rate} & \textbf{Overall Misuse} \\
    \midrule
    Pass@1 &  39.06\% & 76.08\% & 29.72\% \\
    Pass@5 &  21.98\% &  93.79\% &  20.61\% \\
    Pass@10 &  16.51\% &  96.27\% &  15.89\% \\
    \bottomrule
    \end{tabular}%
    }
    \caption{Pass@k results of GPT 3.5 (T=1, one-relevant-shot).}
    \label{tab:pass-k}
\end{table}

\begin{finding}
    Some LLMs can learn from the correct usage example, which reduce the API misuse rate.
\end{finding}

\subsection{Robustness Analysis}
We evaluate the benchmark on GPT 3.5 under different temperatures (Table~\ref{tab:temp}). From the result, changing temperature does not significantly change the misuse rate and compilation rate. To study the effect of different prompting methods, we study how the API misuse rate changes when we replace the one-shot examples with the API usage rules. We feed the symbolized rules to ChatGPT and get the rules in natural language. We add the usage rules as part of the prompts and evaluate GPT-3.5 with \dataset{}. The results are shown in Table~\ref{tab:ret-prompt}, which indicates that the API usage rules might not help reduce the API misuse rate compared to one-shot relevant examples. 
\begin{table}[!htbp]
  \centering
      \small
    \resizebox{\linewidth}{!}{
    \setlength{\tabcolsep}{1mm}{
    \begin{tabular}{lccc}
    \toprule
    \bf Temperature & \textbf{Misuse Rate} & \textbf{Compilation Rate} & \textbf{Overall Misuse} \\
    \midrule
    T = 0 &  38.56\% & 80.71\% & 31.13\% \\
    T = 0.5 & 39.77\% & 80.13\% & 31.87\% \\
    T = 1.0 & 39.06\% & 76.08\% & 29.72\% \\
    \bottomrule
    \end{tabular}%
    }
    }
    \caption{Results of GPT 3.5 with different temperature (Pass@1, one-relevant-shot).}
    \label{tab:temp}
\end{table}
\vspace{-1em}
\begin{table}[!htbp]
  \centering
    \resizebox{\linewidth}{!}{
    \setlength{\tabcolsep}{0.5mm}{
    \begin{tabular}{lccc}
    \toprule
    \bf Prompt & \textbf{Misuse Rate} & \textbf{Compilation Rate} & \textbf{Overall Misuse} \\
    \midrule
    API Usage Rule &  65.01\% & 79.78\% & 51.86\% \\
    One-shot-relevant &  38.56\% & 80.71\% & 31.13\% \\
    \bottomrule
    \end{tabular}%
    }
    }
    \caption{Results of GPT 3.5 with API usage rules (T=0, Pass@1).}
    \label{tab:ret-prompt}
\end{table}

\begin{finding}
    Increasing temperature or replacing one shot examples with API rules does not affect the API misuse rate significantly.
\end{finding}

\subsection{Error Analysis}

\begin{figure}[!htb]
    \centering
    \includegraphics[width=\linewidth]{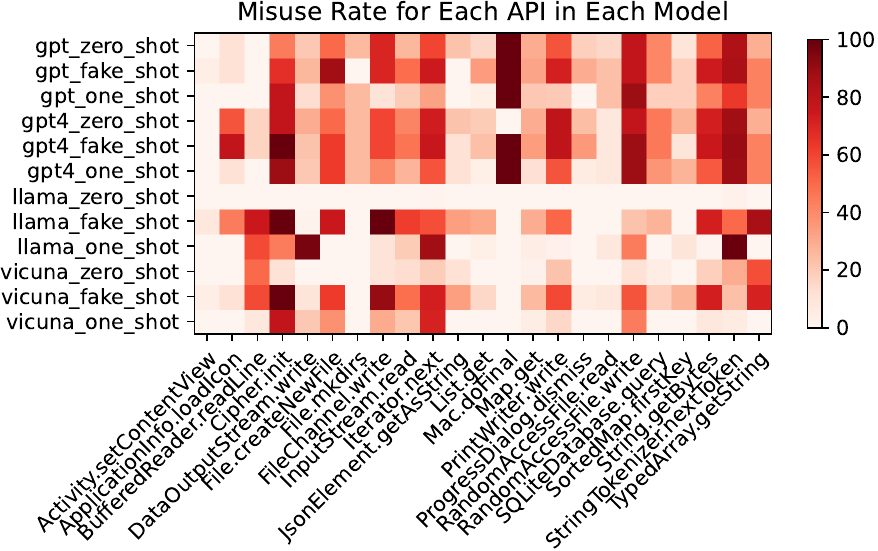}
    \caption{Misuse rate of each API by each LLM. The \textit{deeper} the color, the \textit{higher} the misuse rate. G3.5, G4, LMA, Vic are short for GPT3.5, GPT4, Llama2, Vicuna1.5.}
    \label{fig:api_heatmap}
\end{figure}

In this section, we discuss the answers from LLMs that cannot pass the API usage check in \dataset{} evaluation. There are two categories for failure cases: cases that are not compilable, and cases that are compilable but contain API misuses as shown in Figure~\ref{fig:api_misuse_rate}. We refer to the ability to be compiled successfully as \textit{compilability}. The compilation failure rate is calculated by dividing the number of cases that can be compiled to the total number of cases in the benchmarks. GPT-4 performs the best among all the models regarding compilability, which has less than 10\% of answers that cannot be compiled across all experiment settings. Adding a few shots to prompts helps reduce the compilation failure rate in the evaluation results for all models. 
As for the API misuse rate, we dive deeper into the APIs that LLMs are prone to misuse. Figure~\ref{fig:api_heatmap} details the misuse rate of each API for each LLM. Among all APIs, the Android development API \texttt{Activity.setContentView} has the lowest misuse rate across all the models.  

\subsection{Case Study: API Misuse in GPT-3.5}
Taking GPT-3.5 as an example, we show a typical sample that GPT-3.5 reacts differently under different experiment settings. This question asks the model to help write a string to the file using API \texttt{PrintWriter.write}. Under zero-shot and one-irrelevant-shot settings, the answers differ slightly but both misuse the API by not catching exceptions. After giving the model the correct API usage example, the model learns how to use the API and responds correctly. 

\begin{Verbatim}[frame=single,fontfamily=cmss,commandchars=\\\{\},fontsize=\scriptsize]
\texttt{\textbf{Zero Shot:}}
\texttt{PrintWriter writer = new PrintWriter("f.txt", true);}
\texttt{writer.write("text to append");}
\texttt{writer.close();}
\texttt{\textbf{One Irrelevant Shot:}}
\texttt{String text = "Hello, World!";}
\texttt{PrintWriter writer = new PrintWriter("f.txt", true);}
\texttt{writer.write(text);}
\texttt{writer.close();}
\texttt{\textbf{One Relevant Shot:}}
\texttt{try \{String text = "Hello, World!";}
\texttt{PrintWriter writer = new PrintWriter("f.txt", true);}
\texttt{writer.write(text);}
\texttt{\} catch (IOException e) \{e.printStackTrace();\}}
\end{Verbatim}

\section{Discussion}
\paragraph{Extend to Other Language} \dataset{} focuses on Java API usage since Java is one of the most widely used languages in software development and has a special niche in web and Android ecosystems so that its API misuses may cause more serious problems in real applications. Theoretically, the method proposed in this paper can also be applied to other languages like Python.
\paragraph{Future Work} The API misuse problem proposed in our research can motivate many further research directions. First, how to improve the quality of generated code aside from functionality alignment. To achieve this goal, in-context learning, fine-tuning, and pre-training would be critical to improving existing models. Besides, other online code community like Github could also be a useful resource to evaluate code models, as proposed in a recent work~\cite{jimenez2023swe}. As we believe, evaluating and improving LLMs on the perspective of real-world software development is a demanding and important problem. 
\section{Conclusion}
In this paper, we propose a benchmark \dataset{} to study the API misuse behaviors in code generated by LLMs. From the benchmark results on state-of-the-art models, we find that API misuse widely exists in large language models even when the code is executable and aligned with users' intention. Under different experiment settings, we explore effective methods of benchmarking and improving the API misuse rate of LLMs. To inspire and accelerate future research on this problem, we open source the dataset and benchmark in \url{https://github.com/FloridSleeves/RobustAPI}.
\newpage
\section{Acknowledgments}

The authors sincerely appreciate the reviewers and chairs of the AAAI for their constructive and insightful comments. Their expertise and thorough reviews have significantly contributed to the enhancement of this paper. 

\bibliography{aaai24}


\end{document}